\documentclass{sig-alternate-10pt}
\usepackage[margin=1in]{geometry}

\usepackage{xcolor}
\usepackage{hyperref}
\usepackage{times}
\newcommand{\FDSP}{KEK}
\newcommand{\hitext}[1]{\textcolor{black}{#1}}
\newcommand{\hitextB}[1]{\textcolor{black}{#1}}

\def\ncp{\vspace*{-0.5ex}}

\begin{document}
\pagestyle{empty}

\title{Federated Data Science to Break Down Silos [Vision]}

\author{%
Essam Mansour\\
  \affaddr{Concordia University, Canada}\\
  % \email{essam.mansour@concordia.ca}
\and
Kavitha Srinivas\\
  \affaddr{IBM Research, USA}\\
  % \email{Kavitha.Srinivas@ibm.com}
\and
Katja Hose\\
  \affaddr{Aalborg University, Denmark}\\
  % \email{khose@cs.aau.dk}
}

\maketitle
%!TEX root = ../paper.tex

\begin{abstract}
Similar to Open Data initiatives, data science as a community has launched initiatives for sharing not only data but entire pipelines, derivatives, artifacts, etc. (Open Data Science). 
However, the few efforts that exist focus on the technical part on how to facilitate sharing, conversion, etc. 
This vision paper goes a step further and proposes {\FDSP}, an open federated data science platform that does not only allow for sharing data science pipelines and their (meta)data but also provides methods for efficient search and, in the ideal case, even allows for combining and defining pipelines across platforms in a federated manner. 
In doing so, {\FDSP} addresses the so far neglected challenge of actually finding artifacts that are semantically related and that can be combined to achieve a certain goal. 
\end{abstract}

% \sloppy
% With the growing importance of data science not only as a research discipline but also as a business model, we have witnessed the advent of a number of diverse data science frameworks and ecosystems. Although each of these platforms offers at least some support for collaboration and sharing, scientists remain mostly locked within the same platform. 
%Whereas there already exist initiatives to share data (Open Data), data science as a community needs initiatives for sharing not only data but entire pipelines, derivatives, artifacts, etc., (Open Data Science). 
%To achieve this vision, {\FDSP} poses research opportunities in various areas spanning across federated data management and AI. These research opportunities cover (i) abstracting, capturing, and managing data science artifacts and datasets in decentralized knowledge graphs (KGs), (ii) learning from these KGs, and (iii) automating several aspects of data science including data preparation, authoring pipelines, and insights analysis. 
%!TEX root = ../paper.tex

\section{Introduction}
\label{sec:introduction}

Open Data initiatives have led to the development of Open Data portals that provide machine-readable and structured datasets in topics, such as health, education,  transportation, agriculture, and food. They are driven, for example, by governments, e.g., USA~\cite{USAPoral}, Canada~\cite{CAPoral},  or organizations, such as WHO~\cite{WHO} and WTO~\cite{WTO}, and provide access to thousands of datasets. 
Encouraged by the availability of this data and the FAIR principles~\cite{wilkinson2016fair}, data science projects are increasingly striving at making datasets and related data science experimentation automatically and efficiently findable, accessible, interoperable, and reusable. This includes sharing data science pipelines and derived insights, such as code, notebooks, datasets, and technical papers.

Unfortunately, despite artifacts of experimentation and creation of pipelines becoming increasingly more open, most of the artifacts %associated with derivation and sharing of insights and data science models 
are scattered across various open source repositories, such as GitHub or GitLab. Furthermore, documentation describing the work is available along with code on Jupyter notebooks, blogs in domains, such as Medium, and open repositories of preprints, e.g., ArXiv. 
Recently, we have therefore seen the rise of initiatives and projects, such as Agora~\cite{Agora}, with the goal of providing the foundations of how to technically combine data science pipelines in decentralized and dynamic environments, where data, algorithms, etc. are distributed.
While these projects concentrate on the question \emph{how to technically combine artifacts}, they neglect questions, such as \emph{what artifacts should be combined (across platforms, servers, etc.) to achieve a certain goal} and \emph{how do we find artifacts that are semantically similar or connected}. 
%In this vision paper, we are closing this gap by proposing a federated data science platform, called {\FDSP}~\footnote{\hitext{KEK is the initials of the authors' first name. Kek means "raiser up of the light" in ancient Egypt.}}, which addresses these neglected questions to break down silos in data science (DS). 
In this vision paper, we are closing this gap by proposing a federated data science platform, called {\FDSP}~\footnote{\hitext{\small KEK is the initials of the authors' first name. Kek means "raiser up of the light" in ancient Egypt.}}, which addresses these neglected questions to break down silos in data science (DS). 

Achieving this vision begins with the need to find, combine, and reuse artifacts as they are currently locked away in silos. There is no well-defined way of sharing these artifacts enhanced with semantic descriptions or even general metadata, neither much within a given data science platform and definitely not across multiple platforms. Thus, data scientists cannot automatically find relevant datasets and build a new pipeline on top of related ones since there is no way to identify them. 
As a practical use case and example, let us consider the case of reproducing experimental results of published articles, and analyzing insights driven from datasets.  

\begin{figure*}
\ncp\ncp\ncp\ncp\ncp
  \centering
  \includegraphics[width=0.73\linewidth]{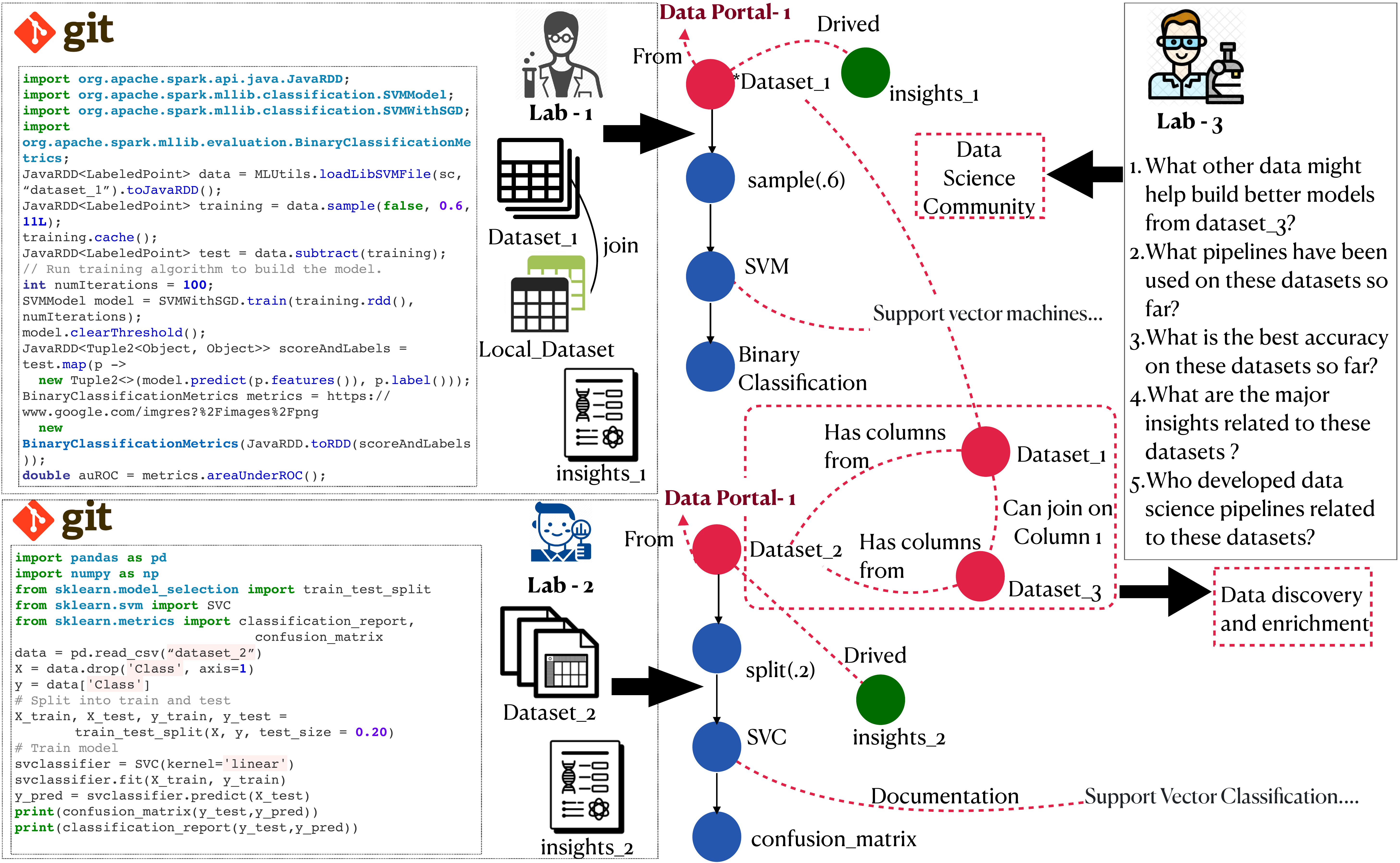}
  \ncp\ncp\ncp\ncp\ncp
  \caption{An overview of data science (DS) experimentations suffering from silos of data, pipelines, and insights. These silos prevent communication among the DS community 
  %to find answers to their questions 
  and lead to consuming more time in data preparation, authoring pipelines, and finding insights related to datasets. The required automation to break down silos is denoted in red color.
  %The dotted red lines and squares denote different aspects demanding automation to break down silos in data science.
}   
  \label{fig:silos}
  \ncp\ncp\ncp\ncp\ncp
\end{figure*}

\textbf{Example. } The problem is illustrated %with an example scenario 
in Figure~\ref{fig:silos} -- Laboratory 1 has a pipeline in a Java-based machine learning library (MLLib) operating on Dataset 1 to produce insights after enriching Dataset 1 with a local dataset; while Laboratory 2 has a pipeline in a Python machine learning library (Sklearn) that operates on Dataset 2 to produce insights described in a recent paper. At a \emph{semantic} level, Dataset 2 could be joined with Datasets 1 and 3. Similarly, the pipelines are \emph{semantically} equivalent; albeit in different programming languages and libraries. Yet, neither laboratory has any way to understand exactly what has been accomplished in the scientific community with respect to the datasets available at a specific data portal, e.g., Data Portal 1.

Existing data science platforms, such as 
MLFlow~\cite{MLflow2O18}
and AutoML~\cite{AutoML}, tend to expand silos by locking-in pipelines and driven insights with limited or no collaboration support to force scientists to use the same platform. While a number of data science portals already exist, such as
OpenML~\cite{OpenML} and Kaggle\cite{kaggle}, they still expect each user to load all open datasets, pipelines, and insights into their specific platforms -- even before users can collaborate. 
% And even though they are quite helpful, the use of any single central platform represents another silo. Although scientists prefer to easily move between platforms based on who they collaborate with and how much they want to share, this is not really supported by existing portals and platforms. In general, 
Access to this community effort should not be restricted to a limited set of APIs, as in Kaggle. %Instead, 
A more flexible mechanism to allow sharing of \textbf{datasets} and their associated \textbf{data science} artifacts is needed.

% Kaggle~\footnote{\url{https://www.kaggle.com/}}

%In this paper we therefore propose a federated data science platform, called {\FDSP}.%\footnote{KEK is the initials of the authors' first name. Kek means "raiser up of the light" in ancient Egypt.} 
%to break down silos in data science (DS). 
{\FDSP} therefore aims to provide a mechanism for the scientific community to discover and learn from each other's work automatically. 
In particular, {\FDSP} will help (i) discover and extract relevant data, (ii) enable scientists to collaborate more effectively regardless of the DS platforms they use, (iii) support efficient discovery of the most recent insights related to a dataset, (iv) enable scientists to reuse and combine (parts of) existing DS pipelines in novel ways, (v) enable reproducibility of experimental results with ease, and (vi) encourage innovative applications to automate several aspects of DS based on the most recent DS experimentation.

One of the key concepts to enable this vision and overcome silos is to abstract from syntactical differences of existing platforms and instead focus on the semantics of datasets, artifacts, and pipelines. Once we understand the semantics, we can more easily identify similar or matching artifacts and combine them %using wrappers or similar technologies to combine them 
in a federated manner. 
Instead of creating yet another silo by limiting {\FDSP} to a non-flexible standard, another key consideration is to retain a maximal degree of flexibility by capturing metadata and semantics in a flexible graph format.
In our example from Figure~\ref{fig:silos}, for instance, each laboratory's artifacts (stored in databases, file systems, or from a GitHub repository) are represented and indexed by an abstract graph representation that can be shared with other laboratories as illustrated in Figure~\ref{fig:overview}. 
%disentangle the platform specific configurations from the semantics of 

%The remainder of this paper is organized as follows. 
We present an architectural overview of {\FDSP} in Section~\ref{sec:platform}. Section~\ref{sec:inUse} discusses how {\FDSP} could be used in practice. We discuss the research gaps for reaching our vision in Section~\ref{sec:opp}, and related work in Section~\ref{sec:rw}. Section~\ref{sec:con} concludes the paper.

%Section~\ref{sec:rw} puts our vision into context with related proposals and the state of the art and Section~\ref{sec:con} concludes the paper.

%!TEX root = ../paper.tex

\ncp\ncp
\section{The {\FDSP} Platform}
\label{sec:platform}

{\FDSP} aims to break up data silos by extracting and representing semantic information about data and artifacts in a flexible graph structure. 
The nature of extraction in {\FDSP} therefore results in a set of labeled graphs that together form decentralized data science knowledge graphs (\textbf{DSKG}s). {\FDSP} manages DSKGs using RDF-based knowledge graph technology because (a) it already includes the formalization of rules and metadata using a controlled vocabulary for the labels in the graphs ensuring interoperability, (b) it has built-in notions of modularity in the form of named graphs, so for instance, each laboratory's specific project could get its own named graph, (c) it is schema-agnostic, allowing the platform to support reasoning and semantic manipulation, e.g., adding new labelled edges between equivalent artifacts, as the platform evolves, and (d) it has a powerful query language with federated support (SPARQL)~\cite{lusail}. %to support federated query processing.

\begin{figure}[t]
\ncp\ncp\ncp\ncp
  \centering
  \includegraphics[width=0.89\linewidth]{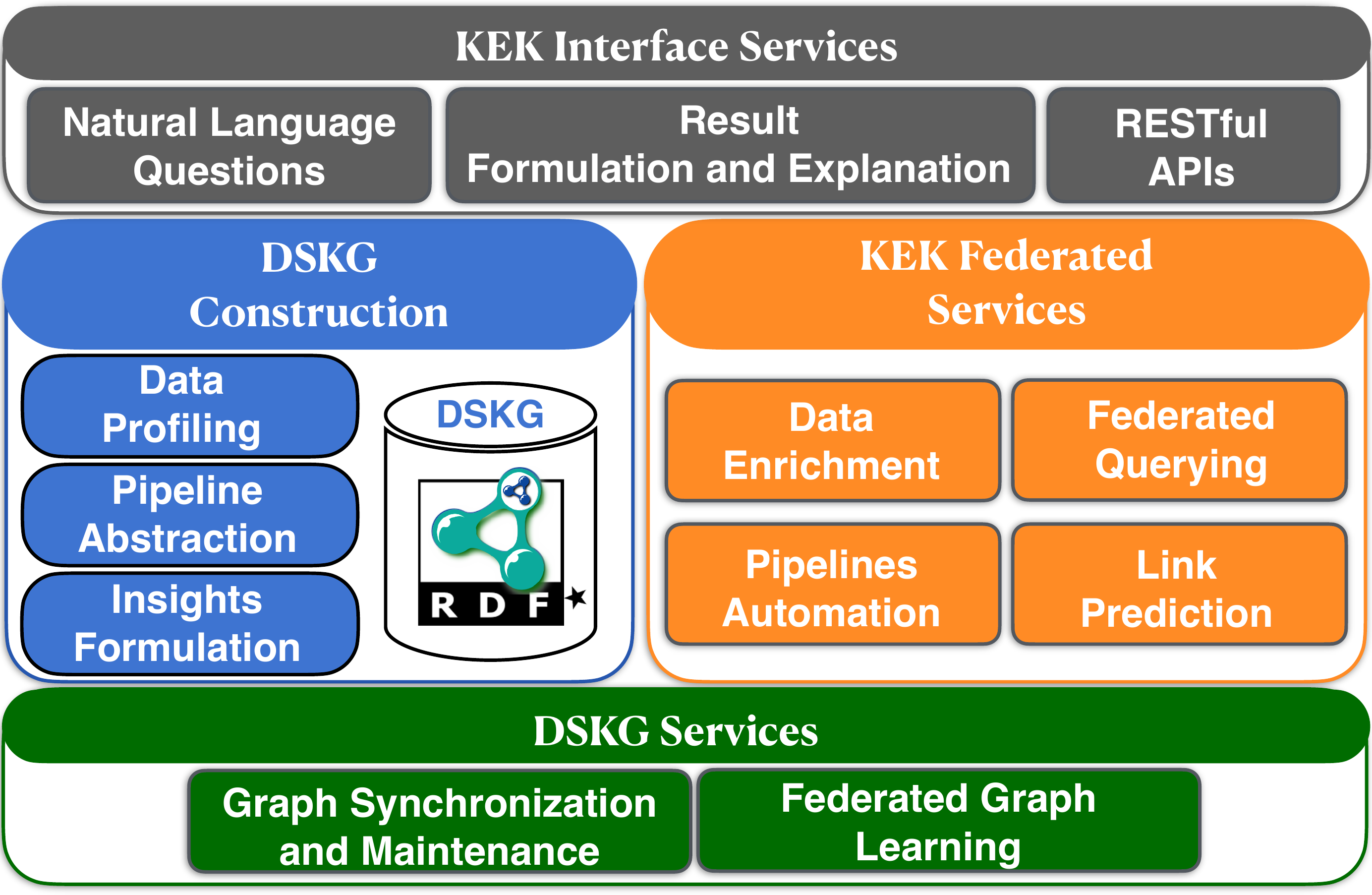}
  \ncp\ncp\ncp\ncp\ncp
  \caption{The {\FDSP} platform architecture.}   
  \label{fig:platform}
  \ncp\ncp\ncp\ncp\ncp\ncp
\end{figure}

% Building upon DSKGs,  
The {\FDSP} platform consists of four main sub-systems, as illustrated in Figure~\ref{fig:platform}, and provides support for federated data science: 
(i) 
extracting semantic information from data items (datasets, pipelines, insights, artifacts, etc.), 
(ii) 
discovering links and similarities among data items at different granularities, such as datasets, tables, and pipelines, 
(iii) 
decoupling the semantics of experimentation on data items (pipelines and insights) from the used data science platform, 
(iv) 
interlinking these semantics with the relevant datasets, 
(v) 
processing complex queries efficiently in geo-distributed settings, 
(vi) synchronize  the  local DSKG with local datasets and scripts of pipelines at scale.

% tracking the vast number of versions (not only dataset versions but also versions of pipelines), and 
% (vii) detecting data/code bias in data science pipelines -- bias probability is high due to the vast pool of connected datasets and pipelines.

\begin{figure}[t]
\ncp\ncp\ncp\ncp
  \centering
  \includegraphics[width=0.75\linewidth]{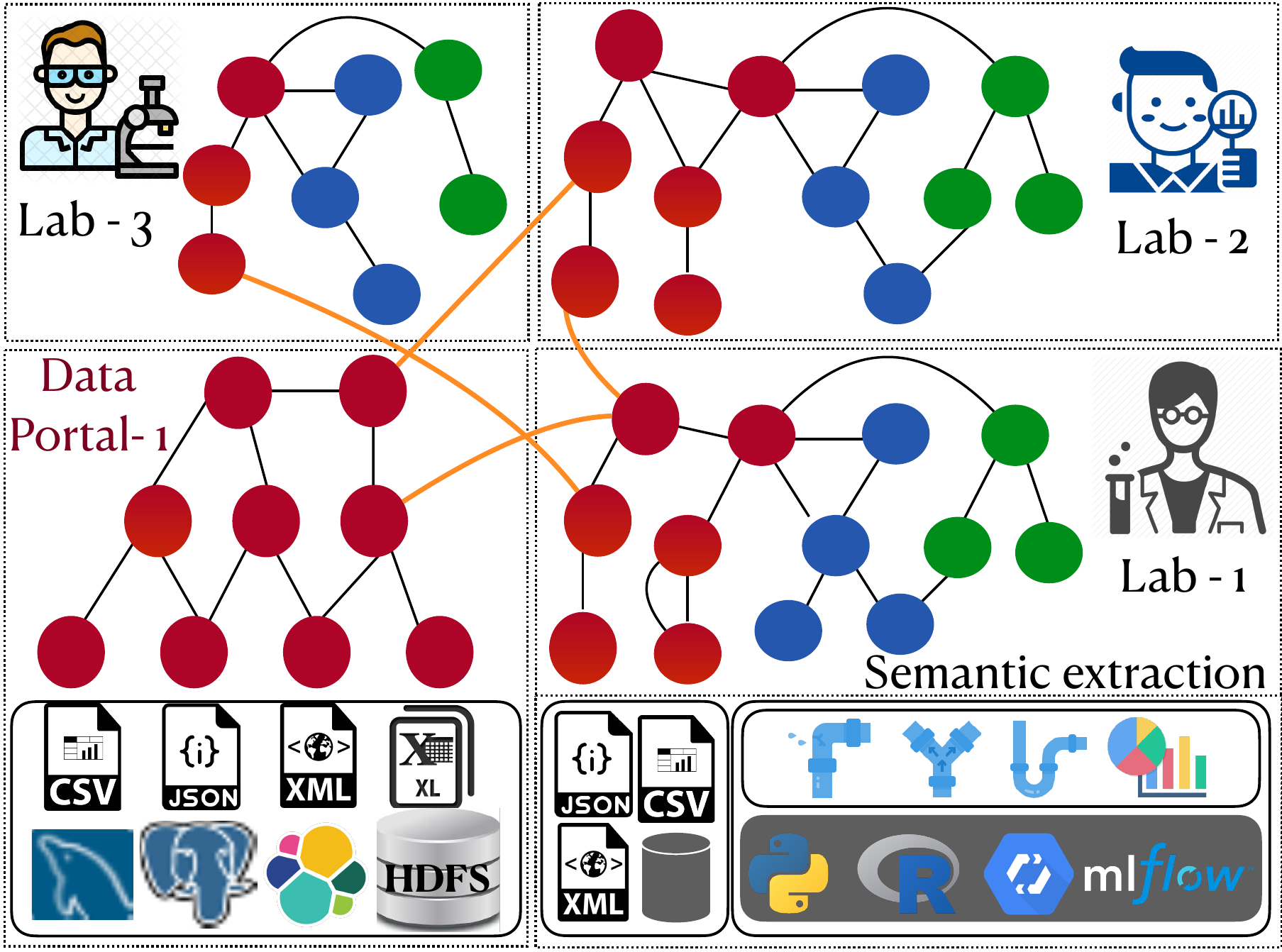}
  \ncp\ncp\ncp\ncp\ncp
  \caption{In {\FDSP}, decentralized data science knowledge graphs interconnect datasets to relevant pipelines and insights.
%   , which are executed and visualized independently by a data science platform. Each DSKG is maintained and managed by an independent {\FDSP} portal. {\FDSP} provides protocols to enable communications among {\FDSP} portals and with the platforms and users.
}   
  \label{fig:overview}
  \ncp\ncp\ncp\ncp\ncp\ncp
\end{figure} 

% \subsection{{\FDSP} Sub-systems}
% Combining these functionalities holds the key for truly federated data science as combining (parts of) pipelines from multiple platforms comes down to finding intermediate pipelines offering wrapper and transformation functionalities to help overcome differences in standards. The {\FDSP} platform consists of four main sub-systems, as illustrated in Figure~\ref{fig:platform}. 

\textbf{DSKG Management. }
In {\FDSP}, the DSKG construction sub-system profiles local datasets to construct a knowledge graph interconnecting data items, e.g., datasets, tables, and columns, accessed locally. The sub-system also maintains DSKG with the semantics captured and extracted from scripts of pipelines and insights. The data owner uses {\FDSP} to publish the graph to be accessible via the Web. In {\FDSP}, the DSKG services index local datasets and pipelines and maintain up-to-date local graphs capturing the extracted semantics. % and support federated graph learning. 

\textbf{{\FDSP} Federated Services. }
{\FDSP} provides federated services over geo-distributed DSKGs to allow automatic discovery and learning from data science projects across multiple data science users and heterogeneous data sources. A key feature of these services is to create and maintain links between decentralized DSKGs via, for example, link prediction. 
Another feature is a query processor that performs federated queries over the local knowledge graph and multiple other {\FDSP} portals to help scientists find and join datasets, pipelines, etc. 
% To support this function, federated services need to support semantic data enrichment and pipeline automation. 

\textbf{{\FDSP} Interface Services. }
\hitext{{\FDSP} is designed to support interoperability with existing data science platforms and enable effective communication with data scientists. Thus, {\FDSP} provides API libraries to enable different data science platforms to communicate with {\FDSP} portals.} In addition to structured queries over DSKGs, {\FDSP} supports natural language questions that help users easily find answers to their questions and extract the required information directly. A {\FDSP} portal is a RESTful server that accepts HTTPS calls.

\textbf{{\FDSP} Foundations. }
To enable automatic learning from DSKGs, {\FDSP} harnesses a broad range of ML approaches including Graph Neural Networks (GNNs)~\cite{Wu2020GNN} to support different functionalities, such as semantic data enrichment and pipeline automation. 
Our vision of {\FDSP} leverages parallelization and computation sharing to efficiently enable analytical workloads. 

%Users can choose to run the {\FDSP} platform on their own servers or use publicly available ones. Different cloud service providers could also provide a {\FDSP} portal as a service with varying degrees of reliability, performance, and security.

% \textbf{KH: depending on who the reviewers are, it might not be sufficient to simply say ``we extract semantics and represent them as a graph'', there's issues regarding ontologies, vocabularies, etc. that an expert might be missing... but a non-expert might not even get the basics of how these things can be captures with knowledge graphs.}{\color{red}{\textbf{EM: I think Kavitha can address this point.}}}\textbf{KS: I decided to just remove RDFS inference because we did discuss the shared vocabularies bit in the next section.}
\hitext{
%!TEX root = ../paper.tex

\ncp\ncp\ncp\ncp\ncp\ncp
\section{{\FDSP} in Use}
\label{sec:inUse}

To avoid \hitextB{the dependency to a central} instance or authority, {\FDSP} is envisioned as a federated platform of independent {\FDSP} portals, as shown in Figure~\ref{fig:overview}. %that collaborate in a federated manner.
Organizations, such as enterprises, countries, or research labs, can then deploy their own instances of a {\FDSP} portal on top of their data lake. {\FDSP} offers a unique way for organizations to maximize data science potentials by capturing and learning from the usage and interdependencies of their data science artifacts \hitextB{including} datasets, pipelines, and derived insights.
Researchers, data scientists, and ML engineers, can deploy a {\FDSP}  portal to capture the semantics of their pipelines and insights and use the {\FDSP} functionality to access artifacts shared by remote {\FDSP} portals.
Hence, the {\FDSP} functionality could be implemented by different systems to run on private or public servers. Moreover, cloud providers can provide {\FDSP} portals as a service with varying degrees of reliability, performance, and security.

\textbf{Bootstrapping. }
When a new {\FDSP} portal wants to join, the first step is to use the \textit{DSKG Construction} component (Section~\ref{subsec:construction}) to analyze the locally available data items, capture provenance, etc. and build a local DSKG covering datasets, processes, pipelines, and insights. The next step is to use the {\FDSP} \textit{Federated Services} (Section~\ref{subsec:federatedServices}) to ``connect'' the local DSKG to the ones from other {\FDSP} portals as illustrated in Figure~\ref{fig:overview}.

\textbf{Maintenance. }
% {\FDSP} is not a static platform. 
As data scientists work on their projects and ideas, new datasets, pipelines, insights, etc., are continuously created. Hence, {\FDSP} portals need to regularly update their DSKG using the Construction components (Section~\ref{subsec:construction}) as well as \textit{DSKG Services} (Section~\ref{subsec:DSKGservices}). Since this naturally affects the relationship to data items at other {\FDSP} portals, the information about the updates are shared, and the DSKG updated using the {\FDSP} Federated Services (Section~\ref{subsec:federatedServices}).

\textbf{Users of the {\FDSP} Platform. }
Different types of users interact with the system in different ways using the {\FDSP} Interface Services (Section~\ref{subsec:interfaceServices}). An administrator, for instance, might need a slightly different interface than a regular user who might prefer to 
% interested in finding similar pipelines. The latter type of users might want to 
 use a natural language interface. % and receive explanations to their query results. 
Executing a user request in general can then easily entail using all other {\FDSP} components illustrated in Figure~\ref{fig:platform}.
%
%\textbf{Use Cases. }
As a concrete example, a researcher might want to work with a new dataset. Using the {\FDSP} infrastructure, it will be possible to find similar or joinable datasets as well as conclusions derived from similar datasets along with the pipelines that were used in the process.
Hence, given a specific task, users can use {\FDSP}  to explore and propose potential analyses that have been used in similar cases.
For data-driven journalism, given some desired insight, the {\FDSP} infrastructure can help find supporting datasets and pipelines.

}
%!TEX root = ../paper.tex

\ncp
\section{Research Challenges}
\label{sec:opp}

This section highlights the open research challenges and opportunities of {\FDSP}'s components.
%This section highlights {\FDSP}'s main components and functionality and discusses the main research challenges and opportunities. We designed the {\FDSP} sub-systems to work as stand-alone systems and integrated among themselves.

%!TEX root = ../paper.tex

\subsection{DSKG Construction}
\label{subsec:construction}

% Pipelines and insights are associated with datasets as depicted in Figure~\ref{fig:silos}. Hence, 
In {\FDSP}, there is a need for novel methods to capture the semantics of a data science pipeline and its driven insights while interlinking the captured semantics with relevant datasets.
% \textbf{DSKG Modeling:} DSKG unifies disparate data science experimentation and helps locate related datasets. We design DSKG as a navigational data structure that interconnects data items, including columns, tables, datasets, pipelines, insights, etc. 
As in other efforts in the search domain (e.g., schema.org) to specify a common vocabulary, one could leverage vocabularies to conceptualize relationships. 
\hitext{Our DSKG includes nodes of different types, such as table, column, function, method, insight,  and pattern. Some examples for edge types are: I) semantic similarity and inclusion dependency to interlink different data nodes, II) flows and reads to interlink code nodes together or to the used data nodes, and III) measure or aggregate to interlink insights related data nodes. We support automated or semi-automated maintenance of vocabularies to retain maximum flexibility.} 

% is an important challenge. 
% By building upon flexible schemas and relationships represented as knowledge graphs in RDF, we retain a maximal degree of flexibility.

% \textbf{KH: when reading this, it is somewhat strange that the some aspects, such as the general vocabulary, ontology, reasoning stuff, doesn't have a name like the others.} {\color{red}{\textbf{EM: I think Kavitha can address this point, too.}}} \textbf{KS: See if this helps.}

\textbf{Data Profiling:}
\label{subsubsec:dataProfiling}
{\FDSP} data profiling aims at breaking down available artifacts into data items (columns, tables, datasets, pipelines, insights, etc.) to identify similarities and relationships. To achieve this goal, we will use the latest state of the art in data profiling and machine learning. 
{\FDSP}, for instance, requires the identification of hierarchies and statistics between data items such that this information can be used to construct a highly interconnected graph representation, in which vertices represent data items while edges represent relationships between them, such as ``similarity".  This graph is further annotated with provenance/metadata information and semantics to arbitrary domains of interest.  

% (e.g., \cite{DBLP:journals/corr/abs-1906-00781})
There is significant work in mapping columns and tables to concepts in knowledge graphs; but much of the work is primarily based on columns with string datatypes.  More recent work has targeted numerical columns (e.g., \cite{DBLP:journals/corr/abs-2012-08594}) but work of this nature is still at a fledgling stage.  
% Another open challenge is doing this at scale on millions of datasets that would be indexed by a system like KEK.
% For example, suppose the schema's of columns A and B are similar. In this case, we represent this information as an edge annotated with the relationship's semantics (similarity and measure), a value representing the degree of similarity, and potentially other information, such as the origin of a data item, e.g., URL of a dataset or file, etc.
% Encoding and querying such information efficiently is an important challenge for the current state of the art that can be addressed by, for instance, using RDF*/SPARQL*~\cite{Hartig19}.
Our DSKGs are deductive graphs that utilize machine learning as well as inference rules to incrementally introduce and enhance the relationships among the different nodes in the graph. Therefore, the local DSKG will eventually be highly interconnected. This helps our profiling and construction process to scale to vast datasets.

\textbf{Pipelines Abstraction:}
Similar programs are written with different APIs and languages. Initial efforts have been made to abstract the semantics of programs using static and dynamic program analysis techniques to extract language-independent representations of data science pipelines~\cite{graph4code,DBLP:journals/pacmpl/CambroneroR19}. Similar efforts capture the provenance of workflows, such as noWorkflow~\cite{10.14778/3137765.3137789}. The example graph in Figure~\ref{fig:silos} (generated using~\cite{graph4code}) illustrates how data flows through specific API pipeline calls, such as \emph{SVM} or \emph{SVC}.  
A key challenge that remains however is how one might recognize similar pipelines across frameworks or languages. \hitextB{There are many aligned benchmarks, such as CodeNet~\cite{CodeNet}, that can be used by statistical models, such as Transcoder~\cite{RoziereLCL20}, to understand similarity across programs.} 
% \hitextB{There are a number of statistical approaches, such as Transcoder~\cite{RoziereLCL20} that learn translations across multiple languages, and many aligned benchmarks, such as CodeNet~\cite{CodeNet} that can be used by statistical models to understand similarity across programs.}  
One could leverage the associated natural language descriptions for APIs (e.g., documentation, forum posts) to generalize across multiple languages and frameworks.
In Figure~\ref{fig:silos}, for instance, the similarity of \emph{SVC} and \emph{SVM} could be derived from text, although this is still clearly an open challenge. Another challenge is to build multi-language independent abstractions for languages, that go beyond abstracting syntax trees.  Systems, such as PROGRAML~\cite{DBLP:journals/corr/abs-2003-10536}, derive abstract program graphs from neural models. These systems show initial promise for the development of language independent abstractions.

% These systems show initial promise for the development of language independent abstractions of programs, which when combined with text can become language independent representations of the semantics of programs.

\textbf{Insights Formulation:}
Data scientists use sophisticated libraries, such as R, Python, or Gnuplot, and tools, such as Tableau, Infogram, or Google Charts, for creating scripts capturing deeper insights from the data. While there are systems that have been proposed for extracting insights from an analysis of the data~\cite{QuickInsights19}, they do not actually mine existing scripts targeting exploratory data analysis (EDA).  Scripts targeting EDA are not easy to search; neither is it straightforward to enable automatic learning on them. There is a need for innovative approaches to capture the semantics of insights from the scripts, combined with comments in the scripts and connect them to their output including insights, observations, etc. Once this is accomplished, derived insights become searchable and processable at scale.

%!TEX root = ../paper.tex

\subsection{{\FDSP} Federated Services}
\label{subsec:federatedServices}

The DSKG Construction analyzes the locally available datasets and scripts to build a local DSKG. The next step is to use the Federated Services to ``connect'' the local DSKG to the ones from other {\FDSP} portals via link prediction, as illustrated in Figure~\ref{fig:overview}. We support federated querying, data enrichment, and pipeline automation on top of the decentralized DSKGs.

% {\FDSP} aims to enable data science pipelines to support the use of geo-distributed datasets and support collaboration in learning and managing these pipelines. Key components to achieving this goal are building upon the current state of the art and cover federated querying, data enrichment, pipeline automation, and link prediction.
% \textbf{KH: semantic data enrichment and link predication almost sounds like it is more part of the previous subsection than this one.} {\color{red}{\textbf{EM: they are related but we separate between the construction phase and the discovery operations. I can clarify this design choice.}}}

\textbf{Link Prediction on DSKGs:}
\label{subsubsec:linkPrediction}
In DSKGs, vertices represent data nodes, such as a node of type dataset, table, or column, or programming nodes, such as classes, functions, or methods, while edges represent relationships between these nodes, such as content similarity or function usage, respectively. \hitextB{We detect links between data items, such as tables or columns, using different methods, such as measuring content similarity. However, there are still other types of nodes or sub-graphs, e.g., a pipeline or insights, where we need to predict links among them. We solve this problem as a link prediction problem for knowledge graph completion using GNN-based models~\cite{zhang2020revisiting,islam2020comparative}. 
{\FDSP} portals work transparently to interconnect different DSKGs and annotate DSKGs with provenance/metadata information.}
In {\FDSP}, learning the embeddings automatically is even more challenging due to the annotations in DSKG, i.e., hyper-relational facts~\cite{GalkinTMUL20}, and the federated setup, which requires developing effective representation learning for datasets and data science artifacts in a geo-distributed environment. 

% {\FDSP} annotates an edge, node, or triple with additional metadata regarding its provenance, confidence, or validity, as described in Section~\ref{subsubsec:dataProfiling}. Hence, we obtain a knowledge graph consisting of hyper-relational facts. Building upon this model represents challenges and opportunities to infer missing connections between data items and artifacts. 

% \textbf{KH: we actually haven't properly defined what the nodes in the graph are.} {\color{red}{\textbf{EM: it is mentioned briefly in the data profiling subsection. I can rephrase that part.}}} 
%Many link prediction methods extract vectorized representations (embeddings) of graph nodes and edges~\cite{zhang2020revisiting,islam2020comparative}. These methods do not efficiently compute the representation of hyper-relational facts~\cite{GalkinTMUL20}. In {\FDSP}, learning the embeddings automatically is challenging due to the federated setup, which requires developing effective representation learning for datasets and data science artifacts in a geo-distributed environment. 

\textbf{Federated Querying and Exploration:}
Building upon knowledge graphs and existing standards, a variety of graph databases, commercial and research prototypes, is already available with basic support of federated querying. 
% Nevertheless, optimizing federated query processing over geo-distributed knowledge graphs remains an ongoing area of active research~\cite{lusail,MontoyaSH17} that covers various aspects of query optimization, including cardinality estimation~\cite{NeumannM11,stefanoni2018estimating}, indexing, exploiting parallelism, cost and response time estimation, non-blocking query operators and incremental execution, etc.
The challenge does not only lie within optimizing query execution across several {\FDSP} portals but also to keep each single one of them responsive despite potentially high query loads. 
% Beyond scalability, another challenge is to keep the system stable and available~\cite{AebeloeMH19}, e.g., by relying on P2P technology and replication. 
Furthermore, {\FDSP} will support fine-grained and non-blocking query execution to produce results progressively. Thus, our federated execution model efficiently enables knowledge graph exploration and supports graph analytics queries generated by components, such as the semantic data enrichment and pipeline automation.

\textbf{Semantic Data Enrichment:}
In the data preparation stage, data scientists tend to generate, in many cases, structured data, e.g., Dataframes, even from data sources of unstructured or semi-structured datasets, such as data logs or  JSON documents. Usually, modeling results show data scientists that there is a need to add supplementary information to enrich the prepared dataset, as these dataframes may cover a limited number of cases.     
{\FDSP} assists users to easily extract relevant data, as discussed in Section 3.4. 
Moreover, {\FDSP} supports semantic data enrichment to find unionable, joinable, combinable data items, discover shortest paths, and schema integration. Users will be able to review discovered data before making the final decision on how to combine and further refine them. %stitching it to the existing dataset. 
{\FDSP} further introduces functionalities to learn from the structure of DSKGs and make automatic recommendations for data enrichment based on semantic and syntactic matching.
% with the prepared dataset. 

\textbf{Pipeline Automation Across Platforms:}
\hitext{
{\FDSP}'s DSKG is able to capture API calls within a program, annotated with function calls and links to the used datasets. \hitextB{For pipelines, {\FDSP} does not join, i.e., combine two pipelines together. Instead, {\FDSP} interlinks similar pipelines to enable automatic graph learning for problems, such as pipeline automation as discussed in~\cite{kgpip}.} 
A DSKG takes the form of a knowledge graph and can be used in combination with deep graph generation networks~\cite{li2018learning} to 
%Hence, we utilize deep graph generation networks~\cite{li2018learning} to 
model and generate pipelines for unseen datasets based on different representation learning techniques~\cite{Wu2020GNN}. Then, we use state-of-the-art hyper-parameter optimization systems, such as FLAML~\cite{flaml} or Auto-SKLearn~\cite{AutoSklearn}, to recommend multiple optimized pipelines, \hitextB{see~\cite{kgpip} for more details.} 
Our model could be used by different ML platforms via {\FDSP} APIs to identify similar datasets to the unseen ones to generate new pipelines. Hence, {\FDSP} will provide a breakthrough for pipeline automation across platforms, i.e., by relying on the DSKGs, to help data scientists build data science pipelines quickly. There is a research opportunity to utilize the relevant datasets and previous analytical tasks to filter and classify generated pipelines.}

%Existing work, such as \cite{DBLP:journals/pacmpl/CambroneroR19}, have used dynamic program analysis to automatically generate pipelines for a new dataset. In {\FDSP}, DSKG is a graph of the API calls within a program, annotated with textual information about each function call and is interlinked with the used datasets. Hence, we employ ML techniques, such as graph neural networks (GNNs)~\cite{Wu2020GNN}, on this large collection of pipelines, to perform all operations useful to data scientists, such as recommend models to be used on a given dataset based on other seen scripts for other similar datasets, understand what sorts of data are incorporated in models for provenance, and so on.  Different ML platforms can use the DSKGs to automate pipelines using AI models. Hence, {\FDSP} will provide a breakthrough for pipeline automation across platforms to help data scientists build data science pipelines easily. 
%It is a research opportunity to utilize the common datasets of interest to a data scientist, previous tasks, and user behaviour to filter and classify the recommended pipelines.

%!TEX root = ../paper.tex

\subsection{DSKG Services}
\label{subsec:DSKGservices}

\textbf{Graph Synchronization:} {\FDSP} is not a static platform. As data scientists work on their projects and ideas, new datasets, pipelines, insights, etc., are continuously created. {\FDSP} platforms need to provide support to synchronize the local DSKG with local datasets and scripts of pipelines. This needs to incrementally maintain the DSKG and support pipelines generated by different platforms. This poses a research opportunity to develop a mechanism that efficiently updates the extracted semantics across scripts generated by different platforms. 

\textbf{Federated Graph Learning:}
{\FDSP} aims at developing a federated graph learning mechanism to learn graph representations (embeddings) across multiple DSKGs. {\FDSP} tasks, such as pipeline automation and semantic enrichment, benefit from this mechanism. We compute local and global features that generate embeddings based on the local and global DSKGs structure and topology. The graph features can be computed via analytical graph queries.
Our federated graph learning is a promising technique to learn directly from the graph structure via sharing nodes' embedding with other remote connected nodes.
This  represents  an  open  challenge for a scale message-passing framework in federated settings, and poses a research opportunity to develop an engine supporting variant embedding techniques for semantic queries~\cite{KGNetDemo}. This engine has to optimize the semantic query execution pipeline, automatically opt for the near-optimal embedding techniques, and estimate the cost of using this specific technique.

%!TEX root = ../paper.tex

\subsection{{\FDSP} Interface Services}
\label{subsec:interfaceServices}

For non-technical users, {\FDSP} provides question answering over DSKGs, automatically decide a data model for formalizing the results, and generate explanations. 

\textbf{Natural Language Questions:}
It is essential to reduce the technical effort required to explore and extract data/code from multiple {\FDSP} portals. 
Mapping a natural language question (NLQ) to a formal query language is challenging due to the ambiguity and multiple interpretations w.r.t. vertices related to data items, pipelines, and insights. 
Existing systems need thousands of annotated questions, such as NSQA~\cite{NSQA}, or require excessive preprocessing, such as such as gAnswer~\cite{gAnswer2018}. 
The preprocessing complexity is proportional to the KG size.

DSKGs are massive decentralized graphs that are frequently updated. Thus, existing systems are impractical as the model should be re-trained from scratch for each update. There is a need for a model incrementally updated or trained independently of the graph.
Thus, there is a need to develop a question answering system trained independently of the DSKG, as demonstrated by KGQAn~\cite{kgqan}. The KGQAn system transforms a question into semantically equivalent SPARQL queries via a three-phase strategy based on natural language models trained generally for understanding and leveraging short English text. 
 \hitextB{This poses a research opportunity to query multiple geo-distributed DSKGs and support natural language code and pipeline search~\cite{feng-etal-2020-codebert}.} 
 %, i.e., a more oriented question answering for data science artifacts.  

%In contrast, we focus on developing NLP-based models for semantic understanding and linking to transform from an NLQ, i.e., a short text, into BGP. Thus, these models will be trained independently of the graph.
% These models segment an NLQ, then estimate the semantic similarity between each segment and a few RDF triples samples to predict the semantically equivalent SPARQL queries on the fly.

\textbf{Results Formulation and Explanation:}
Our methodology will develop different methods to estimate the query results' accuracy and index the NLQ segments and their relevant nodes and edges. The index will enhance the semantic understanding and linking of new NLQs based on the seen queries. The models will help in ranking query results. {\FDSP}'s interface services should support data extraction in different formats based on the context of a given task and the NLQ semantic. For example, a data scientist may look for "Metro stations in Montreal," "Politicians born in New York City," or "Pipelines predicting car accidents in Aalborg". The result is not restricted to only one data model, e.g., a table format in the SQL language. 

The result of these questions could be formalized as a map, table, or control flow graph, respectively. This represents an open challenge for adaptive models to predict the optimal formulation of results, e.g., as a table, graph, or map. Moreover, we need to annotate the results of NLQ with an explanation. Our methodology will adjust the query result's data model based on the NLQ semantics and its relevant data elements. This data model will include data explanations to help a data scientist understand the results in the context of a given task.

%!TEX root = ../paper.tex
\ncp\ncp
\section{Related Work}
\label{sec:rw}
\sloppy

{\FDSP} is an end-to-end platform that enables the data science community to automatically discover, explore, and learn from existing data science artifacts and related datasets. 
\hitext{The vision behind {\FDSP} is independent from or complementary 
%orthogonal 
to systems, such as Agora~\cite{Agora} or Cerebro~\cite{Cerebro21}, which focus on more technical aspects of executing data science pipelines across platforms, such as %, i.e., better utilization and unification of multiple computing resources, and systems managing data as assets for trading, such as DMMS~\cite{FernandezSF20}.}
better utilization and unification of multiple computing resources or managing data as assets for trading, such as DMMS~\cite{FernandezSF20}. 
{\FDSP}, in contrast, is operating on a higher level of abstraction and could be built on top of the technical solutions provided by these systems.
}

In {\FDSP}, scripts of pipelines, and insights are managed by platforms of the user's choice. {\FDSP} captures the semantics of these scripts. Different tools, such as Vizier~\cite{Vizier} and Ursprung~\cite{Ursprung}, support the reproducibility of ML pipelines. The users can utilize these tools to manage their scripts without affecting {\FDSP}. LabBook~\cite{conf/bigdataconf/KandoganRSHTCM15} uses crowd sourcing to create a centralized knowledge graph to manage metadata about people, scripts and datasets, but {\FDSP} automatically extracts connections, in a highly distributed setting.
Auto-Suggest~\cite{Auto-Suggest} is a tool helping in auto-completing a data-preparation pipeline. {\FDSP} focuses on modeling the detected insights and interlinking them with relevant datasets and pipelines. This will help automate several  aspects of data science pipelines. Thus, these tools could benefit from {\FDSP}'s knowledge graphs. 

Systems, such as Google's Dataset Search Engine~\cite{GoogleDataset} and  Helix~\cite{Helix}, enable search over metadata of available datasets. \hitext{Data discovery systems construct navigational data structures in the form of a linkage graph, such Aurum~\cite{Aurum}, an RDF knowledge graph, such as KGLac~\cite{KGLacDemo}, or a hierarchical structure, such as RONIN~\cite{RONIN}. 
\hitextB{Data sketches~\cite{sketches} can identify identical datasets used in different environments but cannot identify semantically similar data items or abstract a pipeline.}
Unlike these systems, {\FDSP} captures and extracts semantics of datasets, pipelines, and insights to construct a knowledge graph for data science enabling better
collaboration in the community.} 

\hitext{Multiple data versioning tools aim to track changes in the data used in ML models to enable reproducibility. Some tools were designed as S3 or Git extensions, such as Quilt~\cite{quilt}, DVC~\cite{DVC}, QRI~\cite{QRI}, DataLad\cite{DataLad}, and Git-LFS~\cite{gitlfs}, to handle large data files. These tools do not handle schema changes, which may lead to breaking the execution of data science pipelines. Model management systems, such as ModelDB~\cite{MODELDB} and MLFlow~\cite{MLflow2O18}, focus on reproducibility and tracing the modeling of experiments by capturing performance metrics, such as hyper-parameter and other values used in training. 
These data/model versioning tools do not capture the semantic abstraction of datasets and data science pipelines as proposed by {\FDSP} to enable advanced discovery and automatic learning.
}

%\hitext{Multiple data versioning tools aim to track changes in the data used in ML models to enable reproducibility. Some of these tools were designed as S3 or Git extensions, such as Quilt, DVC, QRI, DataLad, and Git-LFS, to handle large data files and allow experiments. These tools do not handle schema changes, which may lead to breaking the execution of data science pipelines. Unlike these tools, Apple MLdp~\cite{AgrawalABBGGLMP19} provides versioning control that tracks schema changes. Model management systems, such as ModelDB~\cite{MODELDB} and MLFlow, focus on reproducibility and tracing modeling experiments by capturing performance metrics, such as hyper-parameter and values used during training. Existing data versioning and model management systems do not capture and interlink the semantics structure of datasets and data science pipelines as proposed by {\FDSP} to enable advanced discovery and automatic learning.}
%!TEX root = ../paper.tex
 \ncp
 \section{Conclusion}
\label{sec:con}

{\FDSP} is a paradigm shift for open data science which brings together various communities, encourages more data scientists to share their work, and in doing so breaks down silos. In {\FDSP}, we utilize knowledge graph technologies to decouple the semantics of data science artifacts, e.g., pipelines and insights, from the data science platforms used to create and execute them. 
In doing so, {\FDSP} helps finding semantically similar artifacts and also finding out which artifacts should be combined to achieve a certain goal. 
%{\FDSP} will enable discoveries beyond the original intent of a dataset and its related experimentations. 
The development of {\FDSP} poses numerous open research challenges 
that require innovative methodologies such as learning from decentralized knowledge graphs managed by geo-distributed {\FDSP} portals. \hitextB{In addition, new benchmarks are needed to mimic different workloads in federated data science.} 

% In this vision paper, we have proposed a novel platform called {\FDSP} for federated data science. 
% {\FDSP} advances open data portals to make the data searchable, it is not restricted to searching the metadata but also linking back open datasets to the semantics of relevant data science artifacts. 
% The vision paper will encourage more research efforts to join the development of federated data science to break down silos.  

%  \newpage
 \bibliographystyle{abbrv}
 \small
\bibliography{references}

\end{document}